\documentclass[sigconf, review=false]{acmart}

\usepackage{makecell}
\usepackage{bm}
\usepackage{multirow}
\usepackage{balance}

\AtBeginDocument{%
  \providecommand\BibTeX{{%
    \normalfont B\kern-0.5em{\scshape i\kern-0.25em b}\kern-0.8em\TeX}}}

\copyrightyear{2021}
\acmYear{2021}
\setcopyright{acmcopyright}\acmConference[MM '21]{Proceedings of the 29th ACM International Conference on Multimedia}{October 20--24, 2021}{Virtual Event, China}
\acmBooktitle{Proceedings of the 29th ACM International Conference on Multimedia (MM '21), October 20--24, 2021, Virtual Event, China}
\acmPrice{15.00}
\acmDOI{10.1145/3474085.3475388}
\acmISBN{978-1-4503-8651-7/21/10}

\acmSubmissionID{1292}

\begin{document}

\fancyhead{}

\title{DocTr: Document Image Transformer for Geometric Unwarping and Illumination Correction}


\author{Hao Feng$^1$*, \quad Yuechen Wang$^1$*, \quad Wengang Zhou$^{1,2}$$^{\dagger}$, \quad Jiajun Deng$^1$, \quad Houqiang Li$^{1,2}$$^{\dagger}$}

\makeatletter
\def\authornotetext#1{
	\if@ACM@anonymous\else
	\g@addto@macro\@authornotes{
		\stepcounter{footnote}\footnotetext{#1}}
	\fi}
\makeatother
\authornotetext{The first two authors contribute equally to this work.}
\authornotetext{Corresponding authors.}

\affiliation{%
	\institution{$^1$CAS Key Laboratory of Technology in GIPAS, EEIS Department, \\ University of Science and Technology of China}
	\country{}
}
\affiliation{%
	\institution{$^2$Institute of Artificial Intelligence, Hefei Comprehensive National Science Center}
	\country{}
}
\email{haof@mail.ustc.edu.cn, wyc9725@mail.ustc.edu.cn, zhwg@ustc.edu.cn, dengjj@mail.ustc.edu.cn, lihq@ustc.edu.cn}

\def\authors{Hao Feng, Yuechen Wang, Wengang Zhou, Jiajun Deng, Houqiang Li}

\renewcommand{\shortauthors}{Feng and Wang, et al.}

\begin{abstract}
In this work, we propose a new framework, called Document Image Transformer (DocTr), to address the issue of geometry and illumination distortion of the document images. Specifically, DocTr consists of a geometric unwarping transformer and an illumination correction transformer. By setting a set of learned query embedding, the geometric unwarping transformer captures the global context of the document image by self-attention mechanism and decodes the pixel-wise displacement solution to correct the geometric distortion. After geometric unwarping, our illumination correction transformer further removes the shading artifacts to improve the visual quality and OCR accuracy. 
Extensive evaluations are conducted on several datasets, and superior results are reported against the state-of-the-art methods. Remarkably, our DocTr achieves $20.02\%$ Character Error Rate (CER), a $15\%$ absolute improvement over the state-of-the-art methods. 
Moreover, it also shows high efficiency on running time and parameter count.
Our code and results are available at \textbf{\url{https://github.com/fh2019ustc/DocTr}}.
\end{abstract}

\ccsdesc[500]{Theory of computation~Scheduling algorithms}
\ccsdesc[500]{Computing methodologies~Reconstruction}

\keywords{Transformer, Document Unwarping, Illumination Correction, OCR}

\begin{teaserfigure}
  \includegraphics[width=\textwidth]{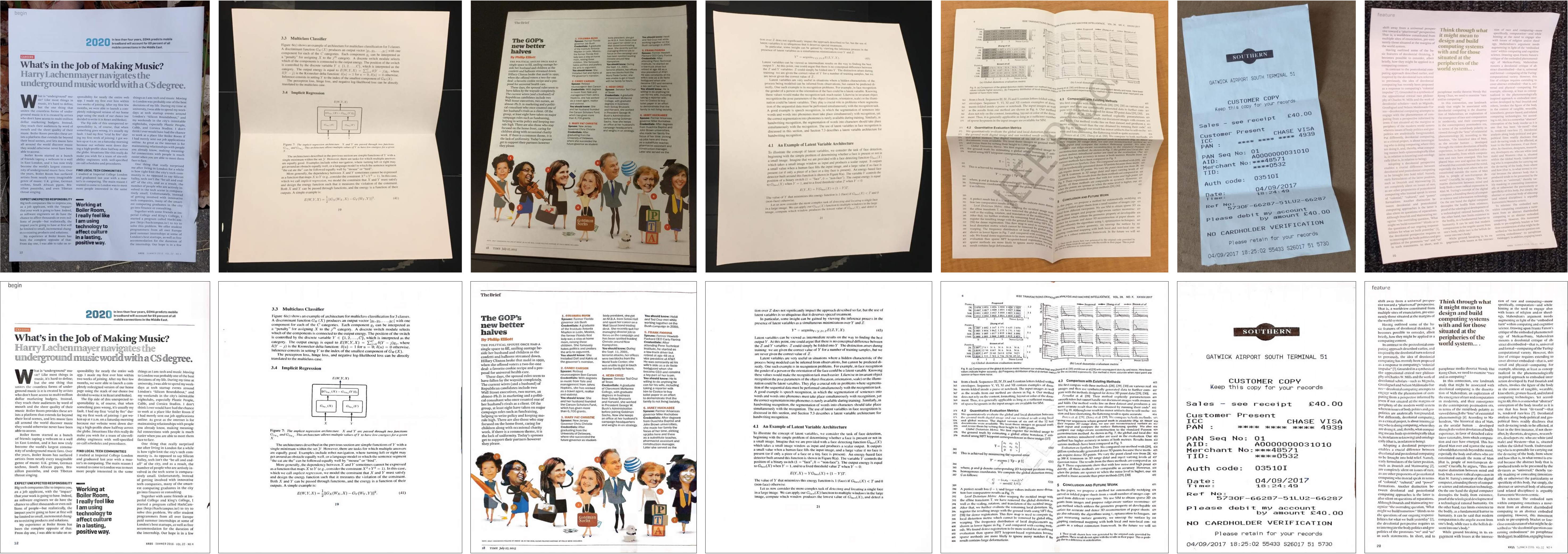}
  \caption{Qualitative rectified results of our Document Image Transformer (DocTr). The top row shows the distorted document images. The second row shows the rectified results after geometric unwarping and illumination correction.}
  \label{fig:teaser}
\end{teaserfigure}

\maketitle

\vspace{-0.04in}
\section{Introduction}
In our daily life, we have to digitize physical documents from time to time. With the advance of smartphone and thanks to its convenience, many people turn to take photos of documents with the camera of smartphones. 
However, such captured images often suffer from uncontrolled geometry and illumination distortion due to various physical deformation of the paper (\emph{i.e.}, folded, curved, or crumpled), camera positions, and uneven illumination conditions. Therefore, they often show bad visual quality and readability. What is more, such distorted document images also cast a negative influence on downstream multimedia applications, such as automatic text recognition, multi-modal retrieval, content analysis, editing, and preservation. This leads to a growing demand for an efficient approach to correct such geometry and illumination distortions. 

Early works on geometric unwarping for document images resort to 3D reconstruction and often rely on auxiliary hardware~\cite{937649, 6909892, 4407722} or multi-view images~\cite{4916075, 7866848} to recover the 3D shape of the paper sheet. However, the involvement of costly hardware or extra shooting requirement unavoidably limits the applicability. Some other methods assume a parametric model on the document surface and optimize the model by extracting specific features such as shading~\cite{1561180}, boundaries~\cite{6628653}, or text lines~\cite{958227,wu2002document}. Besides the non-trivial cost of the optimization process, their problem formulation is oversimplified that leads to sub-optimal performance.

Recently, deep learning has been introduced to address the issue of geometry and illumination distortion, and shown as an alternative to traditional approaches. For geometric unwarping, deep learning methods formulate the task as a pixel-wise displacement regression problem. However, current methods commonly ignore the fact that the physical deformation of the paper is an integral whole in which the parts are interrelated. Specifically, Ma~\emph{et al}.~\cite{8578592} directly regress a dense 2D coordinate mapping field with a stacked UNet~\cite{ronneberger2015u}. Das~\emph{et al}.~\cite{9010747} explicitly model the 3D shape of the paper sheet with a U-structure network. These methods adopt the practice of stacking convolutional modules to perform the regression, while CNN is incapable to capture the long-range relationship to model the deformation of paper. As a result, the rectified documents still exist curved text regions, which carry some important information of the documents.
Additionally, Li~\emph{et al}.~\cite{li2019document} propose to unwarp a distorted document by stitching the unwarped patches. However, it is difficult and ambiguous for a network to understand the original shape of a figure when it is incomplete in a patch image. Hence, it is insufficient to recover the geometry distortion.

For illumination correction, Das~\emph{et al}.~\cite{9010747} regress a low-resolution shading map, which is upsampled to the original size of the high-resolution document. The document image is corrected by matrix multiplication with the high-resolution shading map. However, such an operation will cause blur and artifacts due to the upsampling process. Recently, Li~\emph{et al}.~\cite{li2019document} stack several residual blocks to directly regress the corrected document patches which are then stitched. However, simply stacking the convolutional layers can not capture the global context that is important to model the illumination variation, as reflected by the regions of shadow remaining in the corrected document images.  

To address the aforementioned problems, we propose a new framework called Document Image Transformer~(DocTr) to simultaneously address the issue of geometry and illumination distortions in document images. 
Inspired by recent success of Transformer~\cite{Vaswani2017AttentionIA} in computer vision~\cite{ViT, Carion2020EndtoEndOD, chen2020pretrained}, we integrate the transformer structure into the document image rectification pipeline.
Specifically, DocTr consists of a Geometric Unwarping Transformer and an illumination Correction Transformer, both of which are the transformer encoder-decoder architecture with task-specific heads and tails. Different from the standard transformer, the input of the transformer encoder and decoder are flattened 2D image features and a task-specific learnable embedding, respectively. 
Given a distorted document image, the geometric unwarping transformer first encodes the features of images and decodes the coordinate mapping field for unwarping. By applying the predicted coordinate mapping to the original image, we can obtain the unwarped image.
Then, we crop the unwarped image into patches with a predefined overlap. The patch images are fed into the illumination correction transformer for illumination rectification. 
Finally, the high-resolution document image can be obtained by stitching the corrected patches.

The main contribution of this paper is the proposal of DocTr, which is the first transformer-based framework for document image rectification.
Thanks to the attention mechanism of transformer,  DocTr is able to capture global information for pixel-level geometry and illumination distortion.
Extensive experiments on several datasets, \emph{i.e.,} Doc3D, DRIC, and DocUNet dataset, demonstrate the effectiveness and superiority of our DocTr over the existing state-of-the-art methods on both tasks. 
Notably, on DocUNet benchmark~\cite{8578592}, we achieve significant improvement on OCR results (absolutely $15.32\%$ Character Error Rate (CER) reduced compared to the state-of-the-art method~\cite{9010747}).
Furthermore, our method shows high efficiency on inference time and parameter count.

\begin{figure*}[tbp]
	\begin{center}
		\includegraphics[width=0.86\linewidth]{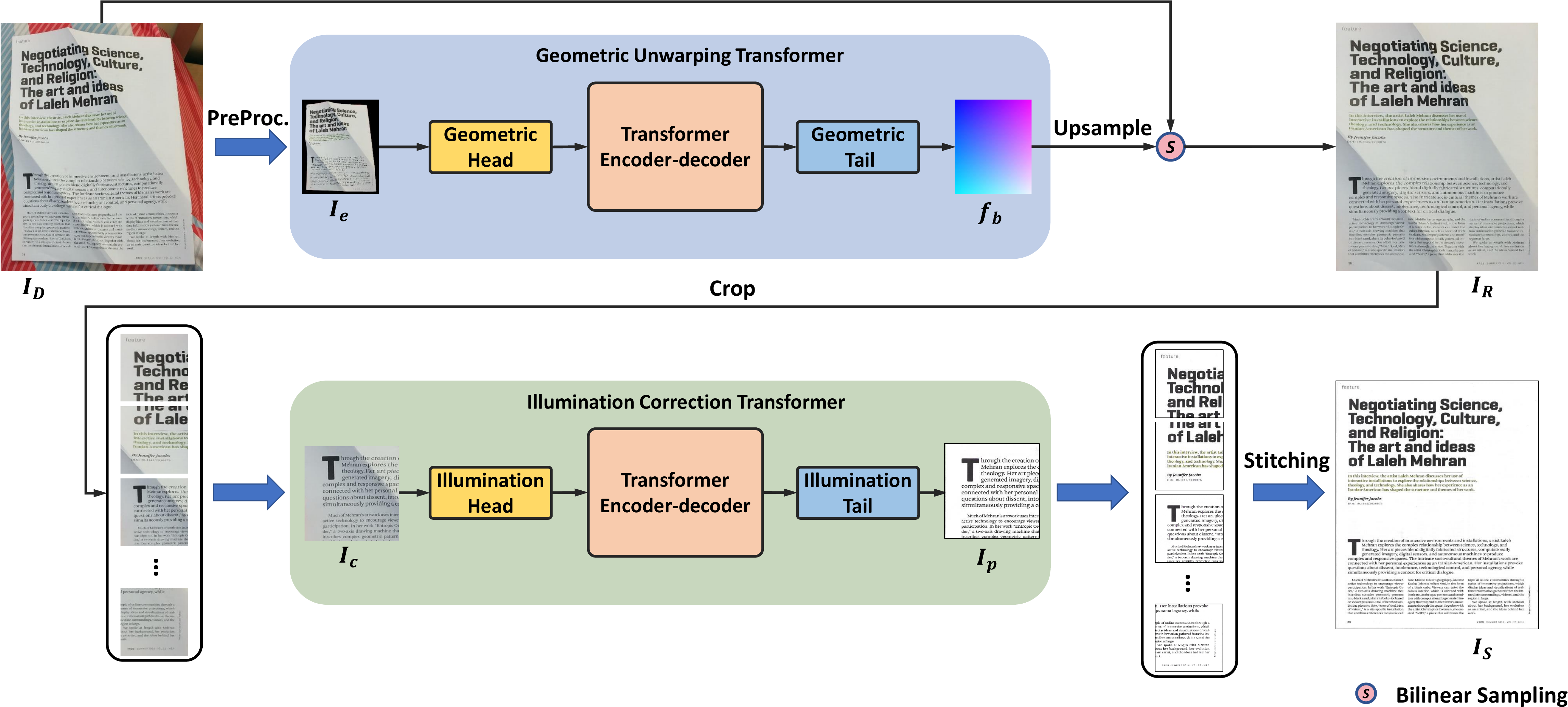}
	\end{center}
	\vspace{-0.1in}
	\caption{
	An overview of Document Image Transformer (DocTr). It consists of two main components: a geometric unwarping transformer and an illumination correction transformer.
	}
	\label{fig:overview}
\end{figure*}

\vspace{-0.1in}
\section{Related Work}

\subsection{Geometric Unwarping}

\textbf{Rectification by 3D shape reconstruction.} Some methods utilize auxiliary hardware to reconstruct the 3D shape of the deformed paper sheet. 
Brown \emph{et al}.~\cite{937649} acquire the 3D representation of shape using a structured light 3D acquisition system. Based on physical modeling technique, Zhang \emph{et al}.~\cite{4407722} use a laser range scanner to perform the modeling. Some other methods utilize multiview images or single image for 3D shape reconstruction. By using multi-view images, You \emph{et al}.~\cite{7866848} propose a ridge-aware 3D reconstruction method, and Tan \emph{et al}.~\cite{1561180} utilize the shape from shading technique for geometric unwarping. Cao \emph{et al}.~\cite{1227630} assume a general cylindrical surface on the surface of document. Das \emph{et al}.~\cite{9010747} explicitly model the 3D shape with a convolutional network.

\textbf{Rectification from low-level features.} The low-level features from a single image also show a useful clue for correcting the geometric distortion. 
In early works, many algorithms aim to correct the curved text lines to be horizontal and straight. The detected text lines are modeled as cubic B-splines by Lavialle \emph{et al}.~\cite{958227}, non-linear curve by Wu and Agam~\cite{wu2002document}, polynomial approximation by Mischke and Luther~\cite{mischke2005document}. 
Ma \emph{et al}.~\cite{8578592} first introduce the neural network to the task and directly regress the pixel-wise displacement with a stacked UNet~\cite{ronneberger2015u}. Li \emph{et al}.~\cite{li2019document} propose to perform the unwarping by stitching the displacement field of the image patches. Xie \emph{et al}.~\cite{xie2020dewarping} estimate pixel-wise displacement using a fully convolutional network with a smooth constraint. Amir \emph{et al}.~\cite{markovitz2020can} localize the words and their orientation in the networks.

\subsection{Illumination Correction}
By feeding a low-resolution document image to a stacked UNet~\cite{ronneberger2015u}, Das \emph{et al}.~\cite{9010747} regress a low-resolution shadow map, typically $256 \times 256$. Then, the shadow map is upsampled to the shape of the original high-resolution document and conducts matrix multiplication with the document image to output the corrected document. This approach is limited by the size of the shadow map as the upsampling process will cause blur. Differently, Li \emph{et al}.~\cite{li2019document} stack several residual blocks to regress the corrected document patches which are then stitched. However, simply stacking the convolutional layers is hard to capture the global context to model the illumination variation, causing the remaining shadow regions on the stitched patches. 
Therefore, we propose a transformer-based network to capture the global information as well as retain local features.

\subsection{Transformer in Language and Vision}
Since it is first proposed by Vaswani \emph{et al}.~\cite{Vaswani2017AttentionIA} for machine translation, transformer has become a prevailing architecture in NLP.
The basic block of transformer is the multi-head attention module, which aggregates information from the whole input in both transformer encoder and decoder module. 
Transformer demonstrates superior performance in language model pretraining methods~\cite{Devlin2019BERTPO, Radford2018ImprovingLU, Yang2019XLNetGA}, and achieves competitive performance on diverse NLP problems.
Recently, transformer has been introduced to various computer vision tasks, such as image classification~\cite{Chen2020GenerativePF}, image generation~\cite{Parmar2018ImageT}, object detection~\cite{Carion2020EndtoEndOD}, semantic segmentation~\cite{wang2021maxdeeplab}, tracking~\cite{Wang2021TransformerMT}, \emph{etc.} Comparing to CNN, the attention mechanism learns more global dependencies, therefore, transformer also shows great performance in low-level tasks~\cite{chen2020pretrained}. Transformer has also been proved effectiveness in multi-modal area, including multi-modal representations~\cite{DeVLBert} and applications~\cite{BotianCaptioning, XinchengTransformer, LiangSemi}. 
Inspired by the extensive applications of transformer, we integrate the transformer encoder-decoder into the document image rectification problem.

\section{Approach}
In this section, we propose a novel framework called DocTr for document unwarping and illumination correction.
As shown in Figure ~\ref{fig:overview}, DocTr consists of a geometric unwarping transformer and an illumination correction transformer.

\subsection{Geometric Unwarping Transformer}
For the geometric unwarping problem, a photographed distorted document image with arbitrary resolution is given as input, the goal is to predict pixel-wise displacement for unwarping.
Specifically, as shown in Figure~\ref{fig:overview}, given an image $\bm{I}_D \in\mathbb{R}^{H\times W\times 3}$, we first downsample it and get the image $\bm{I}_d\in \mathbb{R}^{H_0\times{W_0}\times{C_0}}$, where $H_0=W_0=288$ in our method and $C_0=3$ is the number of RGB channels. 
Then, $\bm{I}_d$ is fed into the preprocessing module to get the background-excluded image $\bm{I}_e$ which is injected into geometric unwarping transformer to predict a backward mapping field $\bm{f}_b=(\bm{f}_b^{u}, \bm{f}_b^{v})$, where $\bm{f}_b^{u}, \bm{f}_b^{v}\in \mathbb{R}^{H_0\times{W_0}}$ denote the horizontal and vertical coordinate mapping map, respectively. 
Finally, we upsample $\bm{f}_b$ to the original size ${H\times{W}}$ of $\bm{I}_D$ and get $\bm{f}_B \in\mathbb{R}^{H\times W\times 2}$. With $\bm{f}_B$, the rectified document image $\bm{I}_R\in\mathbb{R}^{H\times W\times 3}$ can be obtained by warping operation based on the bilinear interpolation as follows,
\begin{equation}\label{equ:task}
	\bm{I}_R(u_0,v_0) = \bm{I}_D(\bm{f}_B^{u}(u_0,v_0), \bm{f}_B^{v}(u_0,v_0)),
\end{equation}
where $(u_{0},v_{0})$ is the pixel position. In the following, we elaborate the key components of our geometric unwarping transformer.

\textbf{Preprocessing.}
Given a downsampled distorted document image $\bm{I}_d \in\mathbb{R}^{H_0\times W_0\times3}$, a light semantic segmentation network~\cite{Qin_2020} is utilized to predict the confidence map of the foreground document, which is further binarized with a threshold $\tau$ to obtain the binary document region mask $\bm{M}_{\bm{I}_d} \in\mathbb{R}^{H_0\times W_0}$. Then, the background of $\bm{I}_d$ can be removed by element-wise matrix multiplication with broadcasting along the channels of $\bm{I}_d$. 
The preprocessing network is trained with a binary cross-entropy loss~\cite{de2005tutorial} as follows,
\setlength{\parskip}{0pt} 	
\begin{equation}
	\mathcal{L}_{seg} = -\sum_{i=1}^{N_p}\left[y_i\log(\hat{p_i})+(1-y_i)\log(1-\hat{p_i})\right],
\end{equation}
where $N_p$ is the number of the pixels, and $y_i$ and $\hat{p_i}$ denote the ground-truth and predicted confidence, respectively.

\textbf{Head.}
Given the preprocessed background-excluded document image $\bm{I}_e \in\mathbb{R}^{H_0\times W_0\times3}$, features are then extracted from $\bm{I}_e$ using a convolutional module $G_{\theta}$ that consists of 6 residual blocks. $G_{\theta}$ downsamples the feature maps at $\frac{1}{2}$ resolution every two blocks and output features $f_g = G_{\theta}(\bm{I}_e)\in\mathbb{R}^{\frac{H_0}{8}\times\frac{W_0}{8}\times c_{g}}$, where we set $c_{g} = 512$.
To adapt to the sequence input form of the subsequent transformers, we flatten $\bm{f}_g$ into a sequence of 2D features, \emph{i.e.}, $\bm{f}_s\in \mathbb{R}^{N_g\times c_{g}}$, where $N_g=\frac{H_0}{8}\times \frac{W_0}{8}$ is the number of patches.

\begin{figure}[tbp]
	\begin{center}
		\includegraphics[width=0.78\linewidth]{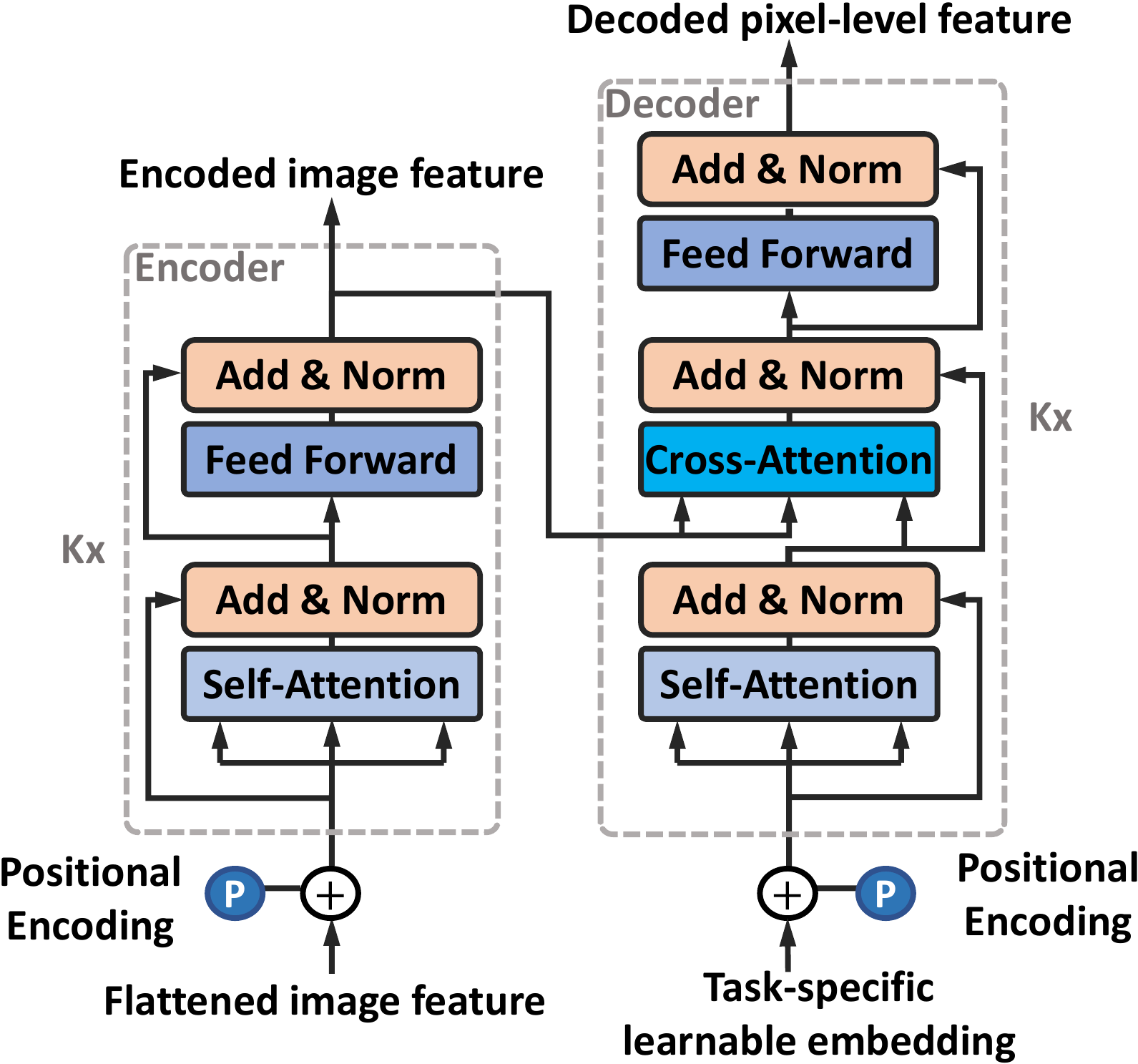}
	\end{center}
	\vspace{-0.1in}
	\caption{Architecture of the transformer encoder-decoder.}
	\label{fig:transstruc}
\vspace{-0.1in}
\end{figure}

\textbf{Transformer Encoder.}
We use a transformer encoder~\cite{Vaswani2017AttentionIA} to encode the global relationship among patches.
During the flattening operation, the flattened image feature $\bm{f}_s$ losses 2D position information, which is essential to learn local and global relationships. Also, since the transformer encoder is permutation-invariant, it is necessary to add position representation explicitly.
Therefore, to maintain the position information in the process, we add learnable 2D position embedding $\bm{E}_p\in \mathbb{R}^{N_g\times c_{g}}$ following \cite{ViT}, which is consistent to different input images.
The resulting 2D position-aware feature $\bm{f}_p = \bm{f}_s + \bm{E}_p$ is then feed into $K$ transformer encoder layers. As shown in Figure ~\ref{fig:transstruc}, each of the encoder layers contains a multi-head self-attention module and a feed-forward network.
The output representation can be calculated as follows,
\begin{equation}
\begin{aligned}
    \bm{F}_0 &= [\bm{f}_{pi}], i\in \{0, 1, \cdots, N_g-1\}, \\
    \bm{Q}_i &= \bm{W}^Q_e\bm{F}_{i-1}, \bm{K}_i = \bm{W}^K_e\bm{F}_{i-1}, \bm{V}_i = \bm{W}^V_e\bm{F}_{i-1}, \\
    \bm{F}^{'}_i &= LN(MA(\bm{Q}_i,\bm{K}_i,\bm{V}_i) + \bm{F}_{i-1}), \\
    \bm{F}_i &= LN(FFN(\bm{F}^{'}_i) + \bm{F}^{'}_i),
\end{aligned}
\label{equ:transenc}
\end{equation}
where $W^Q_e, W^K_e, W^V_e \in \mathbb{R}^{M\times c_{g}\times c_w}$, $M$ is the number of attention heads, $i$ denotes the $i^{th}$ layer of the transformer encoder, $MA(\cdot)$, $FFN(\cdot)$, $LN(\cdot)$ denote the multi-head attention, feed-forward network, and layer normalization, respectively. $\bm{F}_i$ denotes the output feature of the $i^{th}$ encoder layer.

\textbf{Transformer Decoder.}
After obtaining the global-aware representations $\bm{F}_K$, we adopt $K$ transformer decoder layers to generate pixel-level predictions. Each of the decoder layers contains a multi-head self-attention module, an encoder-decoder multi-head attention modules, and a feed-forward network. The encoded feature $\bm{F}_K$ is used as attention value in the encoder-decoder attention computation in each decoder layer.
Different from the original transformer decoder, which performs decoding in an iterative way, we decode the feature of each position in parallel. Therefore, a learnable embedding $\bm{E}_d\in \mathbb{R}^{N_g\times c_g}$ as well as position embedding $\bm{E}_p$ are taken as the input of transformer decoder.
Formally,
\begin{equation}
\begin{aligned}
    \bm{Y}_0 &= [\bm{E}_{pi} + \bm{E}_{di}], i\in \{0, 1, \cdots, N_g-1\}, \\
    \bm{Q}_i &= \bm{W}^Q_d\bm{Y}_{i-1}, \bm{K}_i = \bm{W}^K_d\bm{Y}_{i-1}, \bm{V}_i = \bm{W}^V_d\bm{Y}_{i-1},  \\
    \bm{Y}^{'}_i &= LN(MA(\bm{Q}_i,\bm{K}_i,\bm{V}_i) + \bm{Y}_{i-1}), \\
    \bm{Q}^{'}_i &= \bm{W}^Q_c\bm{Y}^{'}_{i}, \bm{K}^{'}_i = \bm{W}^K_c\bm{F}_K, \bm{V}^{'}_i = \bm{W}^V_c\bm{F}_K, \\
    \bm{Y}^{''}_i &= LN(MA(\bm{Q}^{'}_i,\bm{K}^{'}_i,\bm{V}^{'}_i) + \bm{Y}_{i-1}), \\
    \bm{Y}_i &= LN(FFN(\bm{Y}^{''}_i) + \bm{Y}^{''}_i),
\end{aligned}
\label{equ:transdec}
\end{equation}
where $W^Q_d, W^K_d, W^V_d, W^Q_c, W^K_c, W^V_c \in \mathbb{R}^{M\times c_{g}\times c_w}$.
The final output of the transformer decoder $\bm{Y}_K \in \mathbb{R}^{N_g\times c_{g}}$ is reshaped to $\bm{f_d}\in\mathbb{R}^{\frac{H_0}{8}\times\frac{W_0}{8}\times c_{g}}$ and then used to make pixel-level predictions.

\begin{figure}[tbp]
	\begin{center}
		\includegraphics[width=0.6\linewidth]{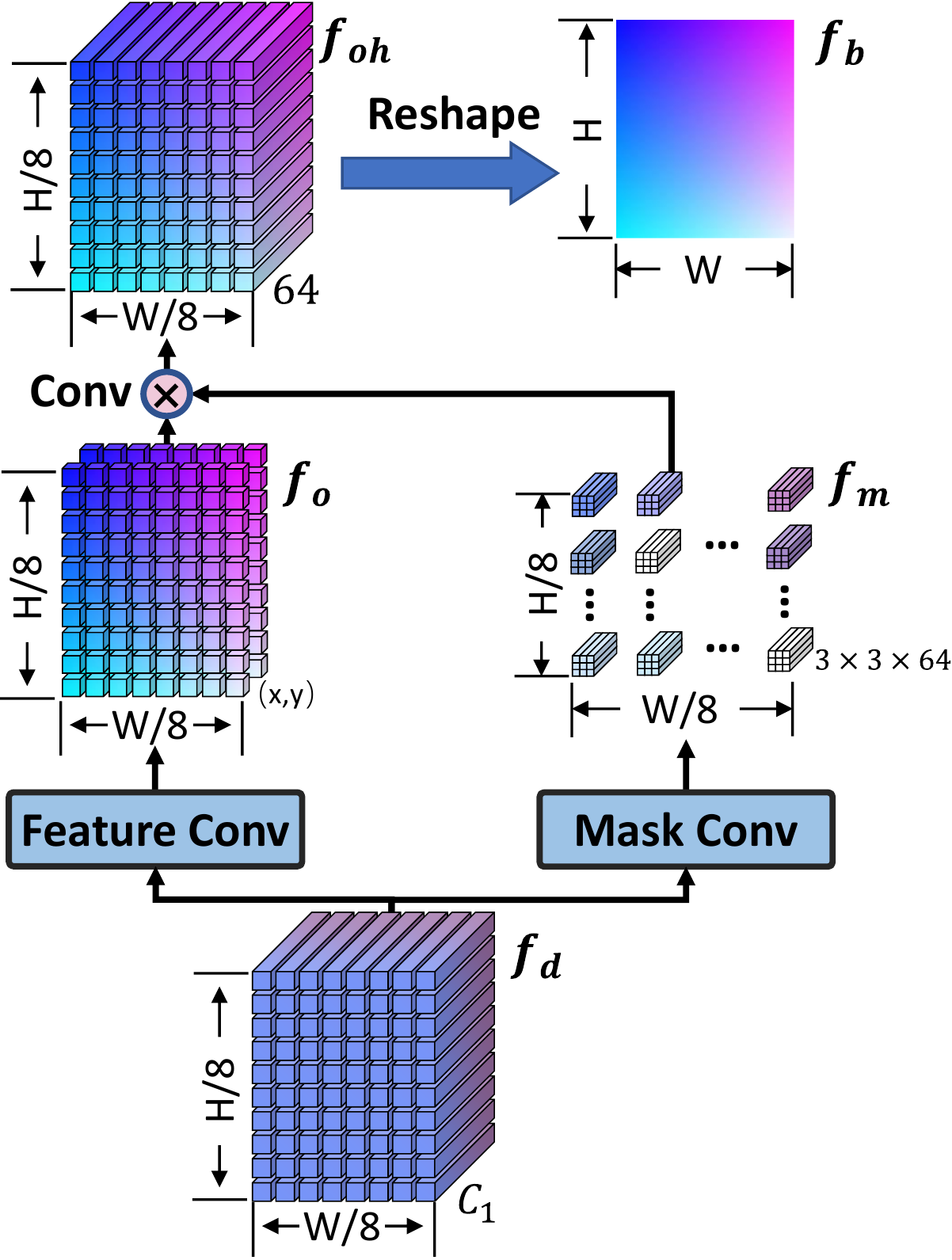}
	\end{center}
	\vspace{-0.1in}
	\caption{Illustration of the tail network for unwarping.}
	\label{fig:GeoTail}
\vspace{-0.2in}
\end{figure}

\textbf{Tail.}
As shown in Fig~\ref{fig:GeoTail}, we train a learnable module to perform upsampling on the decoded features $\bm{f_d}$ and obtain high-resolution predictions. 
Specifically, we first use a two-layer convolutional network to reduce the channel dimension $c_g$ to $2$, \emph{i.e.}, $\bm{f_o}\in\mathbb{R}^{\frac{H_0}{8}\times\frac{W_0}{8}\times2}$, where the two channels represent horizontal and vertical displacement, respectively.
Then, an additional two-layer convolutional network predicts a weight mask $\bm{f}_{m, ij} \in \mathbb{R}^{3 \times 3 \times 64}$ for each pixel $(i, j)$ on $f_d$, where a 2-dimensional softmax is performed on the first two dimensions of $\bm{f}_{m, ij}$. 
Finally, we calculate the convolution of each $3\times3$ patch on $\bm{f}_o$ and the corresponding pixel-wise weight mask. Formally, 
\begin{equation}
\begin{aligned}
    \bm{f}_{oh}^{i, j} &= \bm{f}_o^{i-1:i+1, j-1:j+1} \otimes \bm{f}_{m, ij} \\
        & =  \sum_{u=-1}^{1}\sum_{v=-1}^{1} \bm{f}_o^{i-u, j-v} \bm{f}_{m, ij}^{u+1, v+1},
\end{aligned}
\end{equation}
where $\bm{f}_{oh}^{i, j}\in\mathbb{R}^{2\times64}$.
The high resolution 2-dimensional coordinate mapping map $\bm{f}_b\in\mathbb{R}^{{H_0}\times{W_0}\times2}$ is obtained by reshaping the last dimension of output $\bm{f}_{oh}$ and flattening into $8\times8$ patches.

\textbf{Loss Function.}
The training loss for our geometric unwarping transformer is defined as the $L_1$ distance between the predicted backward mapping $\bm{f}_b$ and ground truth $\bm{f}_{gt}$ as follows,
\begin{equation}
	\mathcal{L}_{geo} = \left \| \bm{f}_{gt} - \bm{f}_b \right \|_1.
\end{equation}

\subsection{Illumination Correction Transformer}
After geometric unwarping, the unwarped document image still suffers from sampling and shading artifacts due to the rectification process and lighting conditions. Hence, we further propose an illumination correction transformer to remove the shadow and improve the OCR accuracy. As shown in Figure~\ref{fig:overview}, given the unwarped image $\bm{I}_R \in\mathbb{R}^{H\times W\times 3}$, we first crop $\bm{I}_R$ into patches with a overlap of $12.5\%$ (each patch at $128\times 128$ resolution). All patches are fed into our illumination correction transformer for illumination correction, and the results are then stitched to get the complete corrected image $\bm{I}_S$. We elaborate correction process in the following.

\textbf{Head.}
Unlike the smooth, continuous backward mapping output of the geometric unwarping network, the illumination correction network outputs a high-frequency image. Hence, we use the architecture without lowering feature resolution. 
Specifically, given a cropped patch image $\bm{I}_c \in\mathbb{R}^{H_p\times W_p\times3}$, we first leverage a 3-layer convolutional module $F_{\theta}$ to extract features $\bm{f}_i = F_{\theta}(\bm{I}_c) \in\mathbb{R}^{H_p\times W_p\times c_i}$. 
Next, we further reshape the patch feature $\bm{f}_i$ into a sequence of flattened $P\times P$ mini-patches, \emph{i.e.}, $\bm{f}_{sp} \in\mathbb{R}^{N_i\times c_i^{'}}$, where $N_i=\frac{H_p\times W_p}{p^2}$ is the number of mini-patches on the input image patch, $c_i^{'} = c_i\times P^2$ is the number of channels of transformer input.
Note that in our method, $c_i$ is set to 16, and $P$ is set to 4.

\textbf{Transformer Encoder-Decoder.}
Similar to the body of the geometric unwarping transformer, we devise an encoder-decoder architecture to encode features of patch images and generate pixel-level illumination correction predictions.
To be concrete, the pre-extracted feature $\bm{f}_{sp}$, added by position embedding $\bm{E}_p^{i}\in \mathbb{R}^{N_i\times c_i^{'}}$, is fed into a K-layer transformer encoder. After being processed by multiple multi-head self-attention and feed-forward layers in the same way as the computation in Equation~\eqref{equ:transenc}, the output feature $\bm{F}_K^{i}$ aggregates global relationship. 
We subsequently use a transformer decoder to decode the illumination distribution and predict corrected pixels. The decoder takes a learnable embedding $\bm{E}_d^{i}\in \mathbb{R}^{N_i\times c_i^{'}}$ and a position embedding as input, and perform multi-head attention to query and aggregate features of different pixels, as in Equation~\eqref{equ:transdec}. 

\textbf{Tail.}
The tail of the illumination correction transformer is a simple convolutional network with one layer. Given the decoded features $\bm{f}_d^i \in\mathbb{R}^{N_i\times c_i^{'}}$ from the decoder, we first reshape it back to the original shape ${H_p\times W_p\times c_i}$ by restoring the 2D mini-patches as well as the image patch. Then, we use a convolutional layer to estimate the corrected patch image $I_p \in\mathbb{R}^{H_p\times W_p\times 3}$. 

\textbf{Loss Function.}
We optimize the illumination correction transformer by minimizing the $L_1$ distance and the VGG loss between the estimated patch image $I_p$ and ground truth image $I_{gt}$ as follows,
\begin{equation}
	\mathcal{L}_{ill} = \left \| \bm{I}_{gt} - \bm{I}_p \right \|_1 + 
	                         \alpha \left \| V(\bm{I}_{gt}) - V(\bm{I}_p) \right \| _1,
\label{equ:lossfun}
\end{equation}
where $\alpha$ is the weight of the VGG loss and $V(\cdot)$ denotes the VGG Network~\cite{simonyan2015deep}. The VGG loss is defined as the $L_1$ distance between the output of the ReLU activation layers of estimated $I_p$ and ground truth image $I_{gt}$. It is based on the pre-trained 19-layer VGG network, and also known as perceptual loss or content loss. 

\section{Dataset}
We evaluate our proposed DocTr on several datasets. To be specific, we train the geometric unwarping transformer on Doc3D dataset, following ~\cite{9010747}. Then, our illumination correction transformer is optimized using the DRIC dataset~\cite{li2019document}. Finally, We evaluate the performance of our DocTr on DocUNet benchmark~\cite{8578592} as previous works~\cite{9010747, li2019document,8578592, xie2020dewarping} suggest. Note that due to the image resolution limit of the Doc3D dataset (\emph{i.e.,} $448\times 448$) which is not suitable for the patch-based illumination correction method, we only use Doc3D for the geometric unwarping task. Besides, the number of document images of DRIC dataset (\emph{i.e.,} 2700) is also not enough for the geometric unwarping task, especially for our transformer-based method. We elaborate individual dataset in the following.  

\textbf{Doc3D.} The Doc3D dataset~\cite{9010747} is the largest dataset to date for the document image rectification task. Created by real document data and rendering software, the dataset consists of 100k distorted document images. For each distorted document image, there are corresponding 3D world coordinate map, albedo map, normals map, depth map, UV map, and backward mapping map.

\textbf{DRIC.} 
The DRIC dataset~\cite{li2019document} consists of 2700 distorted document images, each at $2400\times 1800$ resolution. For each distorted document image, there are corresponding backward mapping map and scanned PDF image. They are also rendered by the rendering software in which many rendering techniques are used. 

\textbf{DocUNet Benchmark.} The DocUNet benchmark~\cite{8578592} is a widely used dataset for document image rectification. It consists of 130 photos of real paper documents captured by mobile cameras. The documents include various types, such as receipts, letters, fliers, magazines, academic papers, and books, \emph{etc}. Besides, their distortion and background are various to cover different levels of difficulty. 

\section{Experiments}
\subsection{Evaluation Metrics}
For geometric unwarping, we use two evaluation schemes based on pixel alignment and Optical Character Recognition (OCR) accuracy. Then, we use image similarity and OCR accuracy to evaluate the performance of illumination correction. To be specific, for pixel alignment, we use Local Distortion (LD)~\cite{7866848} as recommended in~\cite{9010747, 8578592, xie2020dewarping} to evaluate the geometric distortion of rectified images. For image similarity, we use Multi-Scale Structural SIMilarity (MS-SSIM)~\cite{1292216} as previous works~\cite{9010747, 8578592, xie2020dewarping} suggest. For OCR, following~\cite{9010747, 8578592}, we choose Edit Distance (ED)~\cite{levenshtein1966binary} and Character Error Rate (CER) to evaluate the capacity on text recognition.

\textbf{Local Distortion.} Local Distortion (LD)~\cite{7866848} measures the average deformation of each pixel, and is defined as the mean displacement error based on the SIFT flow~\cite{5551153} $(\Delta \bm{x}, \Delta \bm{y})$ from the ground truth scanned image to the rectified image. The SIFT flow is a 2D displacement field that maps each pixel from the scanned image to the rectified image. Then, LD is calculated as the mean value of $L_{2}$ distance between all matched pixels.

\textbf{MS-SSIM.} The Structural SIMilarity (SSIM)~\cite{1284395} measures the similarity between two images and is calculated on various windows of an image. However, the perceived quality of an image usually depends on the sampling density. For this reason, Multi-Scale Structural SIMilarity (MS-SSIM)~\cite{1292216} propose to build a Gaussian pyramid of the two images and is calculated as the weighted summation of SSIM across multiple resolutions. 

\textbf{ED and CER.} Edit Distance (ED)~\cite{levenshtein1966binary} is a way of quantifying how dissimilar two strings are to one another, which is defined as the minimum number of operations required to transform one string into the reference string. The involved edit operations include deletions $(d)$, insertions $(i)$ and substitutions $(s)$. Then, Character Error Rate (CER) can be calculated: $(d+i+s) / N,$ where $N$ is the character number of the reference string. Following~\cite{9010747, 8578592}, we use Tesseract (v3.02.02)~\cite{4376991} as the OCR engine.

\subsection{Implementation Details}
We implement our DocTr in Pytorch~\cite{paszke2017automatic}. We train the geometric unwarping transformer and illumination correction transformer independently on the Doc3D dataset~\cite{9010747} and DRIC dataset~\cite{li2019document}.

\textbf{Geometric Unwarping Transformer.} We first train the preprocessing segmentation module using Doc3D dataset~\cite{9010747}. During training, we randomly replace the background of the distorted document images with the texture images from Describable Texture Dataset~\cite{6909856} for argumentation. 
It is trained for 45 epochs with a batch size of 32. The Adam optimizer~\cite{kingma2014adam} is employed with a learning rate of $1\times10^{-4}$ that is reduced by a factor of 0.1 after 30 epochs. We binarize the confidence map with threshold $\tau$ of 0.5.

Then, we use the Doc3D dataset~\cite{9010747} to train the geometric unwarping transformer. During training, we remove the background of distorted images and resize the images to $288 \times 288$. We use AdamW optimizer~\cite{loshchilov2017decoupled} with a batch size of 8. It is trained for 500k iterations. The learning rate reaches the maximum $1\times10^{-4}$ after 700 iterations and is reduced based on the One-Cycle policy~\cite{smith2019super}.


\textbf{Illumination Correction Transformer.} During training, we randomly crop scanned PDF images and the document images in DRIC dataset [17] which are unwarped by the ground truth backward mapping. The size of the obtained image patches is $128 \times 128$. We train the network for 35 epochs based on the AdamW optimizer~\cite{loshchilov2017decoupled} with a batch size of 24. The initial learning rate is set as $1\times10^{-4}$ and reduced by a factor of 0.3 after 20 epochs. We set the hyperparameter $\alpha=1\times10^{-5}$ in Equation~\eqref{equ:lossfun}.


\subsection{Experimental Results}
In the following experiments, we denote the proposed \underline{Geo}metric Unwarping \underline{Tr}ansformer and \underline{Ill}umination Correction \underline{Tr}ansformer as \textbf{GeoTr} and \textbf{IllTr}, respectively. 

\begin{figure}[t]
\begin{center}
	\includegraphics[width=1\linewidth]{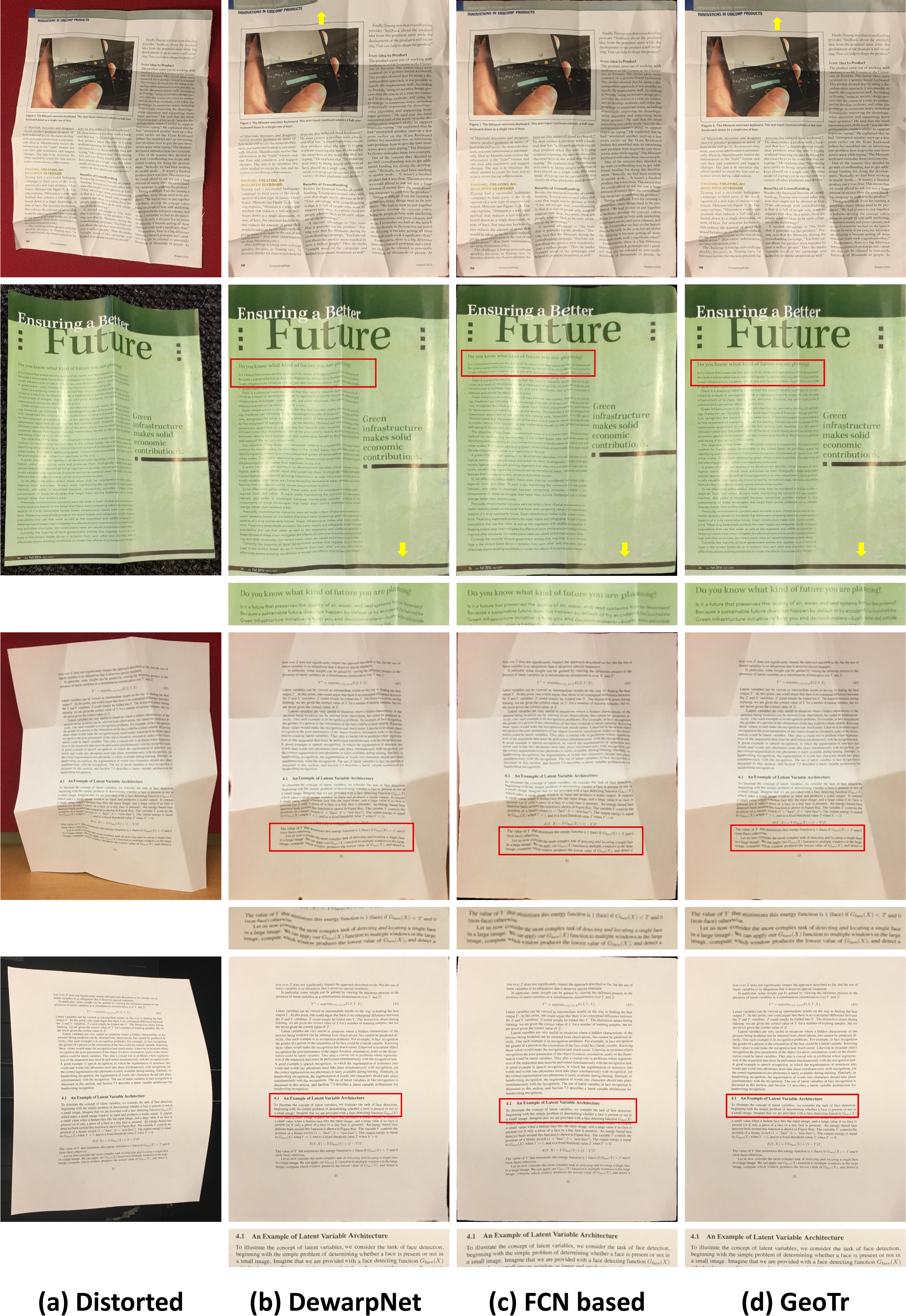}
\end{center}
\caption{\textbf{Qualitative comparison with state-of-the-art methods.} Column 1: original distorted images, column 2: results of DewarpNet~\cite{9010747}, column 3: results of~\cite{xie2020dewarping}, column 4: ours. The comparison of rectified boundaries and text lines are highlighted by yellow arrows and cropped text, respectively.}
\label{fig:geo_com}
\vspace{-0.2in}
\end{figure}

\textbf{Geometric Unwarping.}
We compare our GeoTr with all previous methods on DocUNet benchmark~\cite{8578592} by quantitative and qualitative evaluation. For quantitative evaluation, we evaluate the pixel alignment and OCR accuracy performance. Note that for OCR accuracy evaluation, we select 30 images from the benchmark in which the text makes up the majority of content, following DewarpNet~\cite{9010747}. As shown in Table~\ref{t1}, our GeoTr achieves state-of-the-art performance on all metrics. Additionally, after further illumination correction by our IllTr, we achieve a CER performance of $20.22 \%$, which achieves an absolute improvement of $15.32 \%$ over the state-of-the-art result by DewarpNet~\cite{9010747}. These results demonstrate that our method can effectively rectify the structure and the content of the distorted documents. For qualitative evaluation, we compare the results with state-of-the-art methods as shown in Figure~\ref{fig:geo_com}. As we can see, the rectified text lines of our GeoTr are much more straight. Besides, the incomplete and redundant boundaries phenomenons that existing in other methods are relieved in our GeoTr.

\setlength{\tabcolsep}{7pt}
\begin{table}[t]
    \small
    \centering
    \begin{tabular}{l|cccc}  
    
    \Xhline{2.5\arrayrulewidth}
    	\textbf{Method} &\textbf{LD}$ \downarrow$ &\textbf{MS-SSIM}$ \uparrow$&\textbf{ED}$ \downarrow$ &\textbf{CER}$ \downarrow$\\  
    	
    	\hline\hline
    	Distorted Image  & - & - & 2051.4 & 0.68  \\
    	
    	DocUNet~\cite{8578592}    & 14.08 & - & - & -  \\ 
    	
    	DRIC~\cite{li2019document} & 18.19 & - & 1840.9 & 0.61 \\ 
    	
    	+ DRIC-ILL~\cite{li2019document} & - & 0.2381 & 1547.9 & 0.52 \\  
    	
    	DewarpNet~\cite{9010747} & 8.98 & - & 1121.1 & 0.38 \\  
    	
    	+ DewarpNet-ILL ~\cite{9010747} & - & 0.4735 & 1007.4 & 0.35 \\  
    	
    	FCN based~\cite{xie2020dewarping} & 8.50 & - & 1167.3 & 0.46 \\   
    	
        \hline
    	GeoTr & \textbf{8.38} & - & \textbf{935.2} & \textbf{0.31}  \\ 
        \hline
    	GeoTr + IllTr   & - & \textbf{0.4970}    & \textbf{576.4} & \textbf{0.20}  \\    	
    \Xhline{2.5\arrayrulewidth}
    	
    \end{tabular}
    \\[6pt]
    \caption{Comparison with all previous works on DocUNet benchmark. The suffix ``-ILL'' denotes illumination correction method. ``$\uparrow$'' indicates the higher the better and ``$\downarrow$'' means the opposite.}
    \label{t1}
\vspace{-0.25in}
\end{table}

\setlength{\tabcolsep}{3.4pt}
\renewcommand{\arraystretch}{1.2}
\begin{table}[t]
    \small
	\centering
	\begin{tabular}{l|c|c|c||c|c|c}
	
		\Xhline{2.5\arrayrulewidth}
		\textbf{GEO. Method} & \multicolumn{3}{c||}{DewarpNet~\cite{9010747}} & \multicolumn{3}{c}{GeoTr} \\
	    \hline
		methods & (a) & (b) & (c) & (d) & (e) & (f) \\
		\hline
		$+\ $DewarpNet-ILL~\cite{9010747}  &$\checkmark$    &  &  &$\checkmark$  &  & \\
		$+\ $DRIC-ILL~\cite{li2019document} &&$\checkmark$  &  &  &$\checkmark$  &  \\
		$+\ $IllTr          &   &   &$\checkmark$    &   &  &$\checkmark$  \\
		\hline
		MS-SSIM $\uparrow$  & 0.47  &  0.44  & 0.42  &-& 0.49  & \textbf{0.50} \\
		\hline
		ED $\downarrow$     & 1007.4  &  939.6  & 869.1  &-& 669.8  & \textbf{576.4} \\
		\hline
		CER $\downarrow$   & 0.36  &  0.32  & 0.30  &-& 0.23  & \textbf{0.20} \\
		\Xhline{2.5\arrayrulewidth}
	\end{tabular}
	\\[6pt]
	\caption{Comparison with previous illumination correction methods on DocUNet benchmark. "GEO." denotes the task geometric unwarping. ``$\uparrow$'' indicates the higher the better and ``$\downarrow$'' means the opposite.}
	\label{t2}
\vspace{-0.25in}
\end{table}

\textbf{Illumination Correction.}
We compare our IllTr with two existing methods denoted as "DewarpNet-ILL"~\cite{9010747} and "DRIC-ILL"~\cite{li2019document}. For quantitative evaluation, we first evaluate the three methods by correcting the geometric rectified results of DewarpNet~\cite{9010747}, as shown in method (a), (b) and (c) in Table~\ref{t2}. Then, considering that the code of DewarpNet-ILL~\cite{9010747} is not available, we correct the geometric unwarped results of GeoTr with DRIC-ILL~\cite{li2019document} and our IllTr. The results are shown in method (e) and (f). In both comparisons, our IllTr shows much more performance gain. We further visualize the comparison with DewarpNet-ILL~\cite{9010747} and DRIC-ILL~\cite{li2019document}, as shown in Figure~\ref{fig:geo_com}. Note that we visualize the DRIC-ILL based on the geometric rectified results of our GeoTr, because the geometric unwarping method of DRIC~\cite{li2019document} is incapable to rectify the document images with background. It can be seen that our method shows less blur and shading than the other methods.

\setlength{\tabcolsep}{6pt}
\begin{table}[t]
    \small
	\centering
	\begin{tabular}{l|l|cc}  
		\Xhline{2.5\arrayrulewidth}
		\textbf{Task} & \textbf{Method} & \textbf{Time (s)}&\textbf{Parameters (M)}\\  
		\hline\hline
		\multirow{5}{*}{GEO.} & DocUNet~\cite{8578592} * & 17.5 & 69.1  \\  
		~ & DRIC~\cite{li2019document} & 8.74 & 47.8 \\
        ~ &FCN based~\cite{xie2020dewarping} * & 0.67 & - \\      
		~ &DewarpNet~\cite{9010747} & 0.14 & 86.9 \\
		~ & GeoTr & \textbf{0.13} & \textbf{26.9} \\    
    	\hline\hline
    	\multirow{3}{*}{ILL.} & DewarpNet-ILL~\cite{9010747} & - & - \\
    	~ & DRIC-ILL~\cite{li2019document} & 3.84 & \textbf{0.5} \\
    	~ & IllTr & \textbf{2.79} & 11.6 \\
		\Xhline{2.5\arrayrulewidth}
	\end{tabular}
	\\[4pt]
	\caption{Comparison of running time and model size with previous methods. The task geometric unwarping and illumination correction are denoted as "GEO." and "ILL.", respectively. "*" denotes the reported results in the original paper.}
	\label{t3}
\vspace{-0.28in}
\end{table}

\renewcommand{\arraystretch}{1.2}
\begin{table}[t]
    \small
	\centering
	\begin{tabular}{l|c|c|c|c|c}
		\Xhline{2.5\arrayrulewidth}
		Base model & \multicolumn{5}{c}{GeoTr} \\
	    \hline
		Setting & (a) & (b) & (c) & (d) & \underline{(e)}  \\
		\hline
		$+\ $Preprocessing &  &$\checkmark$  & $\checkmark$  &$\checkmark$  &$\checkmark$   \\
		$+\ $Encoder       &$\checkmark$  &  & $\checkmark$  &$\checkmark$  &$\checkmark$  	\\
		$+\ $Decoder       &$\checkmark$  &$\checkmark$  &   &$\checkmark$  &$\checkmark$   \\
		$+\ $Upsampling Tail &$\checkmark$  &$\checkmark$  &$\checkmark$ &  &$\checkmark$   \\
		$+\ $Bilinear Up.  &   &  &   &$\checkmark$  & \\
		\hline
		MS-SSIM $\uparrow$ & 0.46  & 0.50 & 0.49  & 0.50   & \textbf{0.50}   \\
		\hline
		ED $\downarrow$   & 839.4 & 590.6 & 602.3 & 599.3  & \textbf{576.4}  \\
		\hline
		CER $\downarrow$  & 0.27  & 0.21  & 0.21  & 0.21   & \textbf{0.20}   \\
		\Xhline{2.5\arrayrulewidth}
	\end{tabular}
	\\[4pt]
	\caption{Ablation experiments on GeoTr. The setting used in our final model is underlined. ``$\uparrow$'' indicates the higher the better and ``$\downarrow$'' means the opposite.}
	\label{t4}
\vspace{-0.28in}
\end{table}

\renewcommand{\arraystretch}{1.2}
\begin{table}[t]
    \small
	\centering
	\begin{tabular}{l|c|c|c|c|c}
	
		\Xhline{2.5\arrayrulewidth}
		Base model & \multicolumn{5}{c}{IllTr} \\
	    \hline
		Setting & (a) & \underline{(b)} & (c) & (d)  & (e)\\
		\hline
		$+\ $$64\times 64$     & $\checkmark$ &       &      &  &         \\
        $+\ $$128\times 128$   &     &$\checkmark$  &          &$\checkmark$ &  $\checkmark$\\
        $+\ $$256 \times 256$   &     &     &  $\checkmark$    &    &         \\
		\hline
		$+\ $Encoder  &$\checkmark$  &$\checkmark$  & $\checkmark$   & &$\checkmark$  \\
		$+\ $Decoder  &$\checkmark$  &$\checkmark$  & $\checkmark$   & $\checkmark$&  \\
		\Xhline{1.5\arrayrulewidth}
		MS-SSIM $\uparrow$  & 0.49 & \textbf{0.50}  & 0.50  & 0.50  &0.49  \\
		\hline
		ED $\downarrow$     & 651.5 &  576.4   & \textbf{551.4} & 590.6  & 602.3  \\
		\hline
		CER $\downarrow$    & 0.23 &  0.20 & \textbf{0.19}  &0.21  &0.21  \\
		\Xhline{2.5\arrayrulewidth}
	\end{tabular}
	\\[4pt]
	\caption{Ablation experiments on IllTr. The setting used in our final model is underlined. ``$\uparrow$'' indicates the higher the better and ``$\downarrow$'' means the opposite.}
	\label{t5}
\vspace{-0.28in}
\end{table}

\textbf{Efficiency Comparison.} We compare the inference time and parameter numbers of our GeoTr and IllTr with previous works on processing a 1080P resolution document image. The evaluation is conducted on an NVIDIA GTX 1080Ti GPU. As shown in Table~\ref{t3}, our GeoTr and IllTr both show higher and comparable efficiency, respectively. Note that the running time of DRIC-ILL~\cite{li2019document} is limited by the bilateral mean preprocessing on document images.      

\begin{figure}[tpb]
\begin{center}
	\includegraphics[width=1\linewidth]{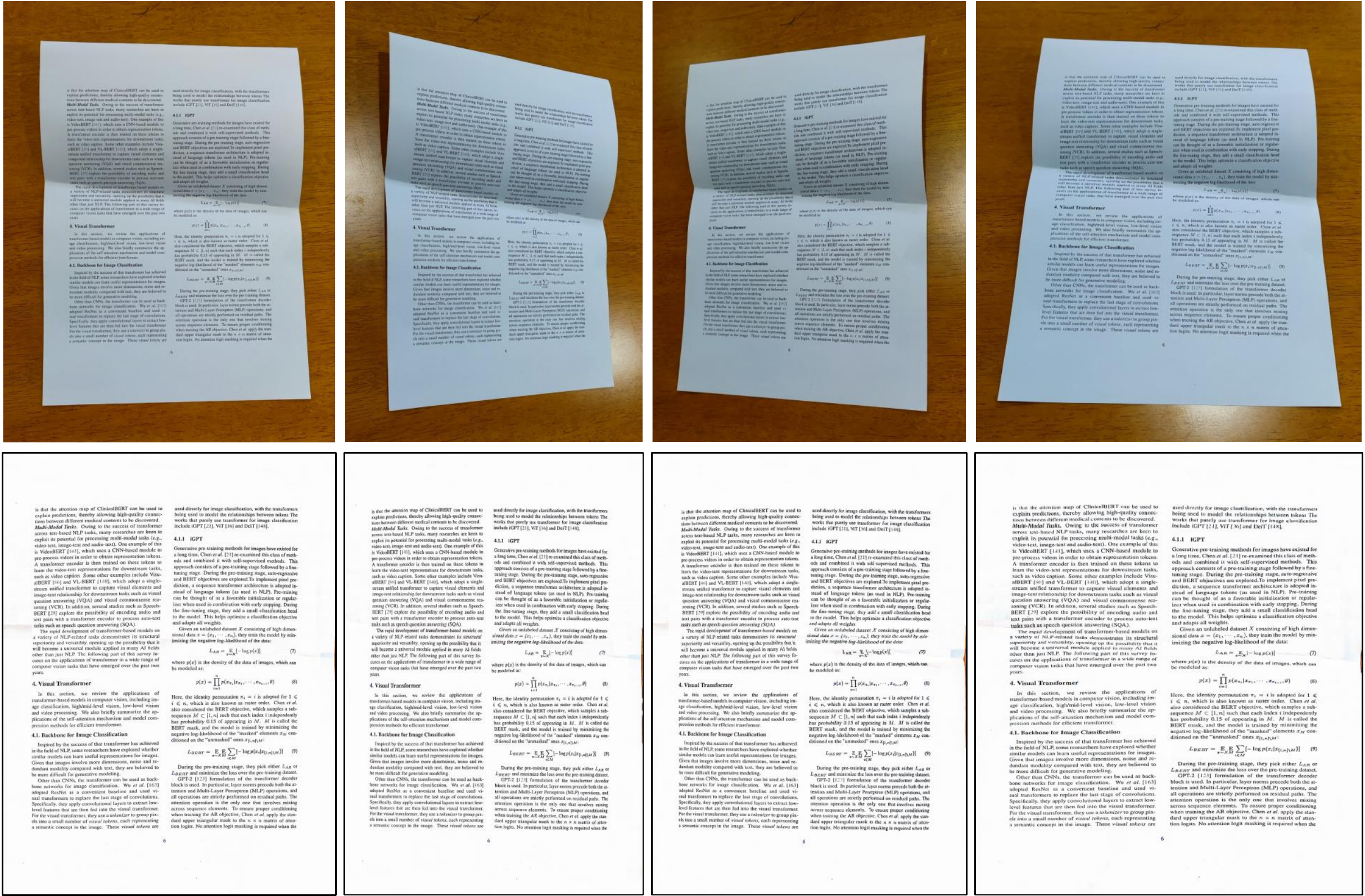}
\end{center}
\vspace{-0.06in}
\caption{\textbf{Robustness of DocTr in terms of viewpoint.} The top row shows the distorted document images taken from different viewpoints. The second row shows the corresponding rectified document images by DocTr.}
\label{fig:robust_view}
\vspace{-0.15in}
\end{figure}

\begin{figure}[tpb]
\begin{center}
	\includegraphics[width=1\linewidth]{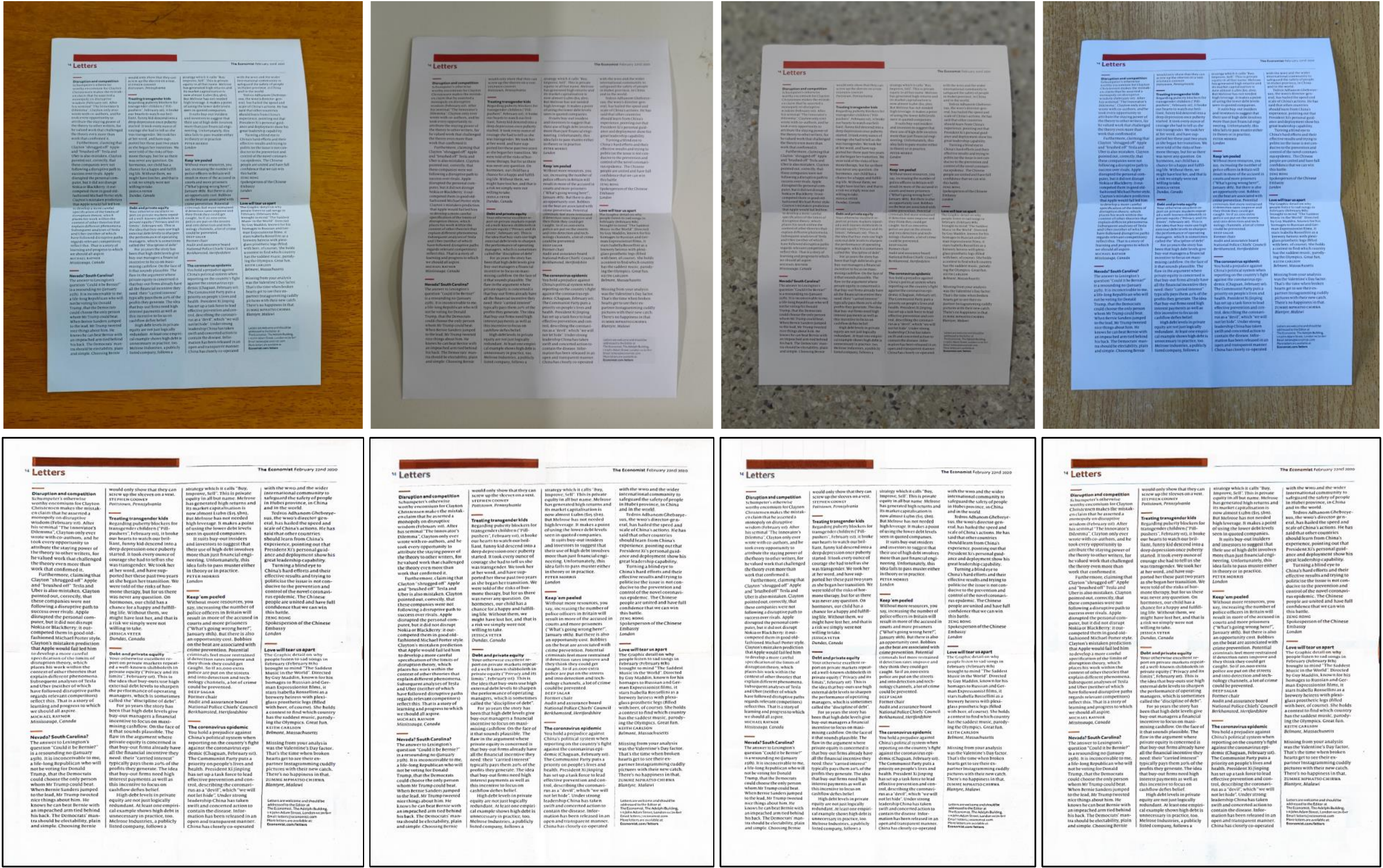}
\end{center}
\vspace{-0.06in}
\caption{\textbf{Robustness of DocTr in terms of illumination variation and background environment.} The top row shows the distorted document images captured in different scenes. The second row shows the rectified document images by DocTr.}
\label{fig:robust_ill}
\vspace{-0.15in}
\end{figure}

\textbf{Robustness.} We further evaluate the robustness of our method in terms of viewpoint and illumination change. The results are shown in Figure~\ref{fig:robust_view} and Figure~\ref{fig:robust_ill}, respectively. It can be seen that our method shows high robustness in various viewpoints, illumination conditions, and background environments.     

\begin{figure}[htpb]
\begin{center}
	\includegraphics[width=1\linewidth]{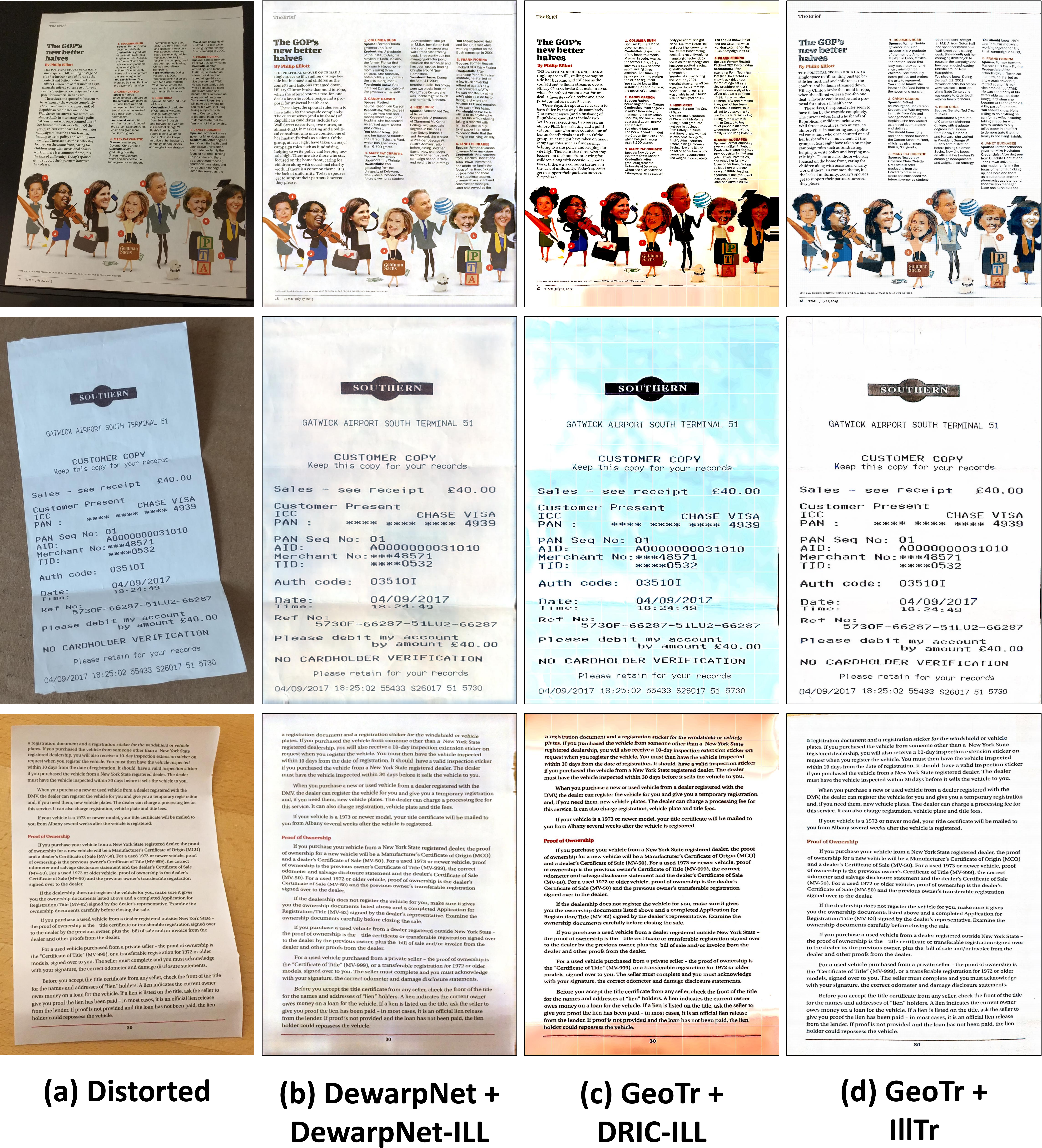}
\end{center}
\vspace{-0.05in}
\caption{Qualitative comparison with illumination correction methods DewarpNet-ILL~\cite{9010747} and DRIC-ILL~\cite{li2019document} on unwarped results of DewarpNet~\cite{9010747} and GeoTr, respectively.}
\label{fig:ill_com}
\vspace{-0.16in}
\end{figure}


\subsection{Ablation Studies}
In this section, we validate the contribution of the key components of our proposed GeoTr and IllTr, respectively. As shown in Table~\ref{t4}, to verify the effectiveness of the preprocessing module, encoder, decoder, and upsampling tail, we evaluate the performance of DocTr under different settings of GeoTr.

\textbf{Preprocessing.} In Table 4, for setting (a) and (e), performance is improved by adding the preprocessing segmentation stage. This is because the various backgrounds of distorted document images involve an extra implicit learning burden for the network to localize the foreground document. As a result, the rectified documents often struggle with incomplete or redundant boundaries, and geometric distortion further spreads the boundary issue to nearby regions. In contrast, given the localized document, the network can focus on the content correction with an integrated document boundary.  

\textbf{Encoder and Decoder.} We independently verify the encoder and decoder module of GeoTr. For setting (b) in Table~\ref{t4}, without any encoder block, we directly feed the flatten image features to the decoder. Then, in setting (c), we feed the encoded image features to the upsampling module without any decoder block. As we can see, our GeoTr (e) achieves better performance than the above two settings. 
This result demonstrates the essential of both the transformer encoder and decoder. Specifically, the self-attention mechanism in encoder block helps to process the global information of the document images. And the decoder, as well as the learned task-specific embeddings, decode the global-aware features into pixel-wise displacement information for tail end prediction. 

\textbf{Upsampling Tail.} The decoder block output features at $\frac{1}{8}$ resolution of input document images. We compare bilinear upsampling with our learned upsampling module. As shown in setting (d) and (e), our upsampling tail produces better results.

For IllTr, we verify the setting of the patch size for stitching, as well as the effectiveness of the encoder and decoder module.

\textbf{Patch size.} We compare the performance of different sizes for stitching. As shown in Table~\ref{t5}, setting (c) shows slight improvement than setting (a) and (b). Following DRIC-ILL~\cite{li2019document}, we use setting (b) (\emph{i.e.}, patch size $128 \times 128$), in our DocTr, to stride a balance between accuracy and inference memory.

\textbf{Encoder and Decoder.} Similar to the experiments for GeoTr, we validate the encoder and decoder module in IllTr. The results are shown in setting (b), (d), and (e) in Table~\ref{t5}. It can be seen that our IllTr (b) outperforms the setting (d) and (e), in which the encoder and decoder blocks are removed, respectively.

\section{Conclusion}
In this work, we present DocTr, a Transformer-based framework, to address the geometry and illumination distortion in document images. Thanks to the processing of global information performed by the self-attention, it achieves state-of-the-art performance on both tasks and significantly relieves the curved text lines and artifacts in rectified results of previous approaches. In addition, it shows comparable or state-of-the-art efficiency. In the future, we will utilize the text content in images to further improve the visual quality and OCR performance.

\begin{acks}
This work was supported in part by the National Natural Science Foundation of China under Contract 61836011 and 61632019, and in part by the Youth Innovation Promotion Association CAS under Grant 2018497. It was also supported by the GPU cluster built by MCC Lab of Information Science and Technology Institution, USTC.
\end{acks}

\bibliographystyle{ACM-Reference-Format}
\balance 
\bibliography{sample-base}


\end{document}